\documentclass[11pt]{article}
\usepackage{geometry}                
\usepackage{graphicx}
\usepackage{amssymb}
\usepackage{epstopdf}
\newcommand{\be}{\begin{eqnarray}}
\newcommand{\ee}{\end{eqnarray}}
\title{The mind as a computational system}
\author{Christoph Adami\\\mbox{}\\
Department of Physics\\University of Bonn}

\date{December 1985}           
\begin{document}      
 \maketitle   
\begin{quotation}
The present document is an excerpt of an essay that I wrote as part of my application material to graduate school in Computer Science (with a focus on Artificial Intelligence), in 1986. I was not invited by any of the schools that received it, so I became a theoretical physicist instead. The essay's full title was ``Some Topics in Philosophy and Computer Science". I am making this text (unchanged from 1985, preserving the typesetting as much as possible) available now in memory of Jerry Fodor, whose writings had influenced me significantly at the time (even though I did not always agree). 
\end{quotation}
\subsection*{The Illusion of Mind}
At the very root of the mind-body problem lies our peculiar perception of what we generally call the mind: our ability to think, to have emotions, pain, motives, fear, to be or not to be in a mood, or to have free will, to name only a few of the notions which occur in the language-game of mind. The perception of this mind as being different from the body has sparked a large number of competing theories of the mind. The present controversy owes a great deal of its sharpness to the fact that Cartesian dualism was shown to be unsustainable, and that materialistic descriptions are in some cases far from convincing. The transformation of the mind-body problem into a mind-brain problem has not changed much on that. Furthermore, the easy escape seems to be barred: indeed, if ever the mind could be shown to be some kind of computing-system, the mind-body problem would instantaneously vanish: the interaction of hardware and software is of course well understood. Unfortunately (and predictably) the attempts to supply machines-with-minds has been as yet unsuccessful, and have drawn critics claiming their failure to be certain out of almost every corner. Hence the statement that the easy escape seems to be barred. It is my aim to argue that this line of thinking is far from being exhausted, and that after it has been shown that in effect there are no obstacles as to adopting this view, the easy escape remains far from easy. 

First I want to concentrate on what seems to be inherent to virtually every theory of mind, which I call the ``Cartesian Premise". 

In part IV of his ``Discourse on Method", Descartes establishes the following~\footnote{Points (1) and (2) are adapted from Flanagan (1984), p. 12}

\begin{quotation}
\noindent (1) I cannot possibly doubt that I exist as a thinking thing.\\
(2) I can doubt, however, that I have a body, and thus that I exist as a physical thing
\end{quotation}

(1) is interpreted as ``One knows for sure that one has a mind, that one is a thinking thing" by Flanagan. He calls this interpretation ``Simple Cartesianism" as opposed to ``Content Cartesianism", ``Causal Cartesianism", and ``Process Cartesianism".\footnote{Flanagan (1984) p. 193.}
I shall adopt the interpretation of Simple Cartesianism. The question that immediately arises is the following: How does one know for {\em sure} that one has a mind?  I can agree with the statement that one knows that one is a thinking thing, if that mans that one {\em notices} that one thinks. I'll agree that one has the impression of having a mind, too. But does the impression of having a mind allow you to infer actually 
{\em having} one? Yes, but only in the case that mind is effectively identified with ``having the impression of having a mind". There is {\em no} way we can rule out the possibility that we actually are subject to the illusion of having a mind. 
The extension of the perspective to the mind nevertheless introduces a highly non-trivial complication: If we are said to have the illusion of mind, then this illusion has to be an illusion {\em somewhere}. It seems as if the only place where we can have illusions is the mind itself! It is however no surprise that we arrive to this kind of self-reference. Indeed, our perception of the mind mainly stems from autointrospection, that is our impression of the mind is our impression of our {\em own} mind. Hence the process of autointrospection is intrinsically self-referential. So we seem to arrive at the conclusion that an impression of mind is only possible when we {\em presuppose} a mind, or else we shall have to believe that we can have illusions in illusions. I shall however propose another way to cope with the illusion of mind. Before setting the stage for this task, I want to come back briefly to the Cartesian Premise. 

I hope that I have convinced the reader that one {\em can} doubt that one has a mind, that it is conceivable that what we call our mind is only an illusion. To follow Wittgenstein however, {\em one cannot reasonably doubt that one has a body}.\footnote{Wittgenstein (1969)} One may argue that all we can have is the impression of having a body, still, to have the impression of intelligence or to have the impression of mind is qualitatively different from having the impression of a body: nobody can touch or see intelligence or mind, therefore the distinction between ``having a body" and ``having the impression of having a body" may (on that level) not be called metaphysical. This would mean that the Cartesian Premise has to be modified. It should now read:
\begin{quotation}
\noindent 1. One cannot reasonably doubt that one has a body.\\
 2. One cannot doubt that one has the impression of having a mind.
\end{quotation}
 I feel I must admit that the original Cartesian Premise was much more poetic. 
 
Before I can turn to the consequences the modification of the Cartesian Premise has on the possibility of AI, I still have to point out how it may be possible to have the illusion of having a mind. 

The aim of Part I was to uncover the metaphysical nature of the distinction between real mind and the illusion of mind. For this discovery  to have any {\em relevance} however, the possibility of the illusion of mind still has to be proved. A convincing proof should in my opinion consist in showing that the propositional attitude of having-the-impression-of-having-a-mind fits into the general description of the mind without leading to inconsistencies. Such descriptions of minds are in general attempts to explain how the mind works without making use of the vocabulary of psychology. Since I shall make no exception to this and since the methodology of reducing a science is a highly intricate problem, I shall outline a framework of reduction that allows me to demonstrate the relevance and validity of the process I will outline subsequently. The basic problems with reduction were already pointed out in Jerry Fodor's essay ``Computation and Reduction".\footnote{Fodor (1981b)} I shall ``rederive" some of his results in my framework and then use the developed terminology for ``proving" the relevance of the reduction of Psychology to Computer Science. 

It may seem at times that the machinery I shall develop to this end is a bit like deploying an artillery to kill a fly. Nevertheless although the exposition of the framework of reduction might be slightly laborious, we shall be rewarded with a powerful terminology. In addition it seems to me that the fly I want to kill is only the tip of an iceberg, and that no less than the arguments developed will do.

 \subsection*{On Reduction}
 Reduction of the sentences $x_i$ of a theory $T_1$ to sentences $y_i$ of theory $T_0$ consists mainly in stating a number of relations between the systems of sentences, and in that effect creating substitution rules. However, a number of constraints have to be applied on this relation, in order for the reduction to meet our intuitive conception of what reduction should be like. The concept seems to be modeled on the cases of successful reduction in the history of science. Some paradigms are the reduction of Chemistry to Electrodynamics and Quantum Mechanics, of Celestial Mechanics to the theory of Gravity, or, in a very recent and from the point of view of reduction highly interesting case, the reduction of the theory of elementary particles to so-called String-theory, as well as the reduction of Gravity to String--theory.\footnote{for a review see \underline{Physics Today}, July 1985 p. 17} It will in this context turn out to be more interesting to study cases where reduction has not been that successful, like this seems to be the case with the reduction of Computational Psychology\footnote{Computational psychology proposes ``computational" accounts for mental processes. It has explanatory power by being able to explain the content of mental states through inference.} to Neurology. 
 
Let $x_i$ be the sentences of a theory $T_1$, $y_i$ the sentences of a theory $T_0$. I shall subsequently write $x_i\in T_1$; $y_i\in T_0$. 
 
The sentences $x$ are built from the {\em vocabulary} $V_1$, elements of the vocabulary $V_1$ shall be called $A_i$; the sentences $y$ are built from the vocabulary $V_0$, $B_i\in V_0$.  
\be
x_i=\sum_{j=1}^{N_1} c_{ij} A_j \ \ \ \ \ \ \ \ i=1,...,n_1\\
y_i=\sum_{j=1}^{N_0}c_{ij}B_j\ \ \ \ \ \ \ \ i=1,...,n_0
 \ee
The vocabulary forms the {\em basis} of the theory, whereas the ``words" linking the elements of the vocabulary, the $c_{ij}$, are the same for both theories. It should be noted that the expressions (1) and (2) should not be taken too literally. The essence is, that sentences in a theory are some sort of ``linear combination" of elements of the vocabulary. Not all {\em possible} combinations though are in effect sentences of the theory. 

I shall now proceed with a number of definitions:
\vskip 0.25cm
\underline{Def. 1}:
\vskip 0.25cm
A {\em Reduction} $R$ is the set of all functions $f_i$, such that
\be y_i=f(x_1,...,x_{n_1})\ \ \ \ y_i\in T_0, \ \ x_i\in T_1, \ \ i=1,...,n_0
\ee

The set of all functions $f_i$ (sometimes called {\em bridge laws}) effectively reduces $T_1$ to $T_0$. I shall call $T_1$ the reduced and $T_0$ the reducing theory. I shall also call the set of all $x_i$ the {\em domain} of $f$: $D(f)$, and the set of all $y=f(x)$ the {\em range} of $f$; $R(f)$. It seems obvious that $R(f)$ is a subset of $T_0$: $R(f) \subset T_0$. 

Let me now introduce some  basic notions and definitions of Algebra to study the properties of $f$:\footnote{see for example Lang (1970) p. 83ff} 
 \vskip 0.25cm
\underline{Def. 2}:
\vskip 0.25cm
A relation $f$ is called a {\em map}, if:

for every $x_i\in D(f)$ there is a $y\in R(f)$ such that $y=f(x)$, {\em and}\\ $f(x_i)\neq f(x_j)\Rightarrow x_i\neq x_j\;.$

This means that for a function to be a map every element from $D(f)$ must be mapped to exactly one element of $R(f)$. I shall give an example for illustration:
\begin{figure}[htbp] 
   \centering
   \includegraphics[width=2in]{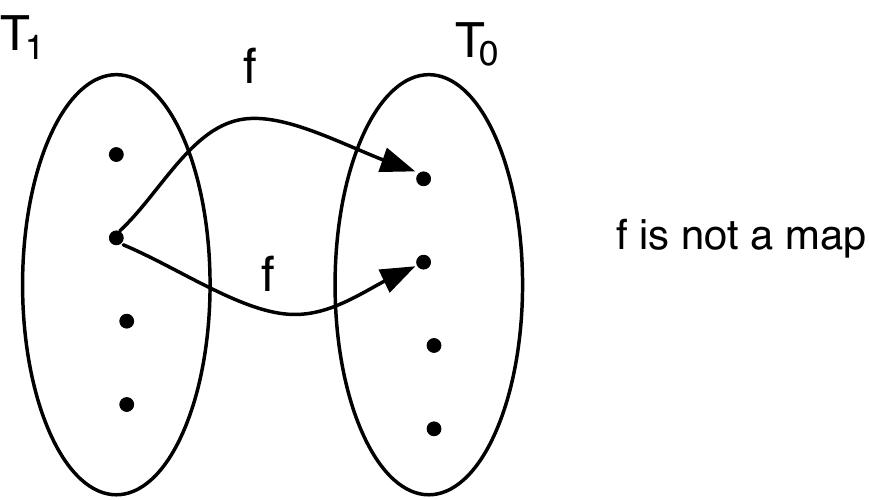} 
   \label{fig:example}
\end{figure}
\vskip 0.25cm
\underline{Def. 3}:
\vskip 0.25cm
A mapping is called {\em one-to-one} (injective) if:

For every $x_i\in D(f)$ and for every $f(x_i)\in R(f): f(x_i)=f(x_j)\Rightarrow x_i=x_j$.
\vskip 0.25cm
This states that if a map is one-to-one, no two elements in $D(f)$ are mapped onto one element of $R(f)$.
\vskip 0.25cm
{\em Example}:

\begin{figure}[htbp] 
   \centering
   \includegraphics[width=2in]{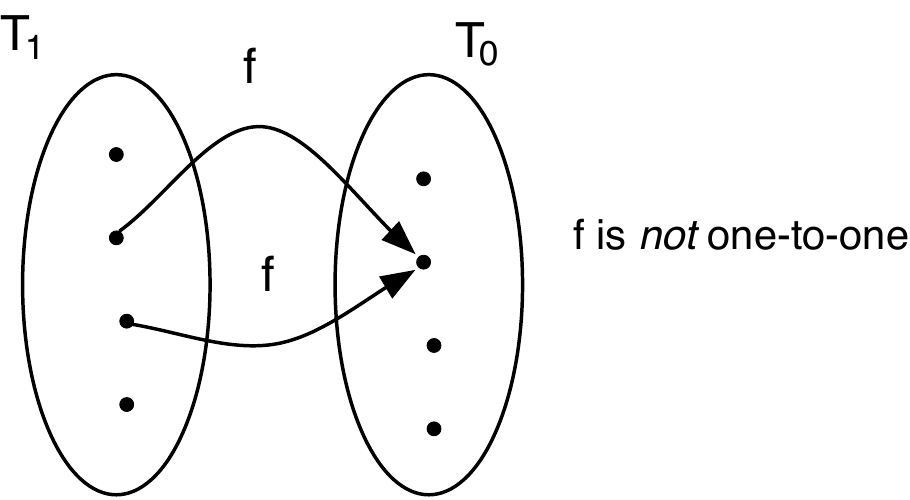} 
   \label{fig:example}
\end{figure} 

\underline{Def. 4}:
\vskip 0.25cm
A map $f$ is said to be {\em onto} (surjective), if:

For every $y\in R(f)$ there exists an $x\in D(f)$ such that: $y=f(x)$. The meaning of this definition is obvious. 
 \vskip 0.25cm
\underline{Def. 5}:
\vskip 0.25cm
A map being simultaneously {\em onto} and {\em one-to-one} is called {\em bijective}. 

Intuitively, a Reduction has to satisfy the following conditions:

\begin{quotation}
\noindent - every element of $T_1$ must have exactly one image in $T_0$;\\
- there must be elements of $T_0$ which have no correspondent in $T_1$.
\end{quotation}
These requirements can now be stated for the Reduction $R$, which is the set of all functions $f_i$:\footnote{It is important to notice that these requirements do not apply to the functions linking the vocabularies, since it is crucial here that any $f_i$ may take more than one argument (see 4b). If the  $f_i$ would take only one argument. the properties of $f$ would carry through to the functions linking the vocabularies. Such a relation is typically given by:
\be A_i=r^{(1)}(B_i) \ \ \ \ A_i\in V_1, B_i\in V_0 \nonumber \\
B_i=r^{(0)}(A_i) \nonumber
\ee
and $f(x_j)=f(\sum_{k=1}^{N-1} c_{jk}A_k)=\sum_{k=1}c_{jk}f(A_k)=\sum_{k=1}^{N_1}c_{jk}B_k=y_j\Rightarrow r^{(0)}=f$
}
\begin{quotation}
\noindent - $R$ must be a map   \ \ \ \ \ \  \ \ \ \ \   \ \ \ \ \ \ \ \ \ \ \ \ \ \ \ \ \ \ \ \ \  \ \                 (4a)\\
- $R$ may or may not be one-to-one       \ \ \ \ \ \ \  \ \ \ \ \ \ \     (4b)\\
- $R$ may not be onto.    \ \ \ \ \ \ \  \ \ \ \ \ \ \      \ \ \ \ \ \ \  \ \ \ \ \ \ \  \ \ \     (4c)     
\end{quotation}

I shall now give some definitions which closely follow Fodor:\footnote{Fodor (1981b) p. 150ff}
   \vskip 0.25cm
  \underline{Def. 6}:
  
  A {\em generalization} $g_i$ from the set of generalizations $G$ is a statement about an event. I shall call $g$ a generalization of $T$ if it is expressed in the vocabulary $V$ but is {\em not} a sentence of $T$:
  \be
  g_i=\sum_j c_{ij} V_j, \ \ \ \ V_j\in V \nonumber
  \ee
  \setcounter{equation}{4}
    \underline{Def. 7}:
    
    A generalization is set to be {\em explained} in $T_1$, if there exists exactly one finite subset of sentences $T_1$ from which generalizations $g$ may be inferred. This fact shall be denoted by 
    \be
    g=\prod_{j=1} ^{\tilde n} x_j\ \ \ \ \ \tilde n\leq n_1
    \ee
  
  I shall give an example as illustration:
  
  \noindent Let $x_1=$ ``All elephants are gray"\\
  \mbox{}\ \ \ \ \  $x_2=$ ``Clyde is an elephant"\\
  
  Then the generalization $g=$ ``Clyde is gray" may be written as
  
  \be g=x_1\otimes x_2=\prod_{i=1}^2 x_i \nonumber
  \ee
  
  The symbol $\prod$ abbreviated the multiple use of  $\otimes$, which simply denotes the logical connection of $x_1$ and $x_2$.\footnote{The nature of the connection as well as the domain of $g$ may be specified. A formulation including these features is in preparation.}
  
  If $g=\prod_{i=1}^2 x_i$ and $x_1.x_2\in T_!$, then $g$ is said to be {\em explained in} $T_1$. 
  
  A generalization may of  course have explanations in different theories, e.g., there may be $x_i\in T_1$ and $y_i\in T_0$ such that:
  \be
  g=\prod_{i=1}^m y_i=\prod_{j=1}^e x_j\;\;. \nonumber
  \ee
  The sets $x_i$ and $y_j$ may then be viewed as different bases to the set of all generalizations, $G$. 
  
  From this point of view a Reduction is nothing but a change of basis for $G$. 
  \vskip 0.25cm
   \underline{Def. 8}:
    \vskip 0.25cm
  A generalization $g\in G$ is called {\em nomologically necessary in} $T$ if it has an explanation  in $T$, that is:
  
  \be
  g=\prod_j x_j, \ \ \ \ \ x_j\in T  \; .
  \ee
  
   \vskip 0.25cm
   \underline{Def. 9}:
   \vskip 0.25cm
A theory $T$ is said to possess {\em explanatory power} if the generalizations of $G$ in $T$ have explanations is $T$, ie.:
\vskip 0.25cm
For every $g_i\in G$ with $g_i=\sum_{ij} c_{ij}A_j$: \ $g=\prod_{i=1}^n x_i\;$.
\vskip 0.25cm
From definitions 8 and 9 it may be inferred that:
\vskip 0.25cm
A theory $T$ has explanatory power if the generalizations of $G$ in $T$ are nomologically necessary.   
\vskip 0.25cm
Therefore, if $T_1$ possesses explanatory power, the reduced theory $T_0$ will necessarily have explanatory power if the Reduction preserves nomological necessity.\footnote{I am not sure whether one may replace the ``if" in definitions 8 and 9 by an ``if-and-only-if" $\equiv$ ``iff", that is, whether the conditions are not only necessary but also sufficient. The question is whether such a definition is sensible.}
  \vskip 0.25cm
   \underline{Def. 10}:
   \vskip 0.25cm
   If a generalization $g$ may be expressed either in the vocabulary $V_1$ or in the vocabulary $V_0$, the substitution rules
   \be
   A_i=r_i^{(1)}(B_1,...,B_{n_0}) \nonumber \\
   B_j=r_j^{(0)}(A_1,...,A_{n_1})\nonumber
   \ee
  define the {\em standard Reduction} of $T_1$ to $T_0$. 

  \vskip 0.25cm
   \underline{Def. 11}:
\vskip 0.25cm   
   A standard Reduction which preserves nomological necessity is called a {\em strong Reduction}. 
  \vskip 0.25cm 
   A Reduction $R$ preserves nomological necessity iff:
 \vskip 0.25cm  
   For every $g\in G$ there are $f_j\in R$ such that: \\
   $
   \prod_{j=1}^m y_j=\prod_{j=1}^m f_j(x_1,...,x_{n_1})=g$, provided that $g=\prod_{i=1}^n x_i$. 
  \vskip 0.25cm 
   Since one is only interested in reduction schemes where the reduced science {\em necessarily} possesses explanatory power, we now have to investigate under what circumstances the Reduction preserves nomological necessity.
   
   For that matter, consider the following diagram:
   
   \begin{figure}[htbp] 
      \centering
      \includegraphics[width=2in]{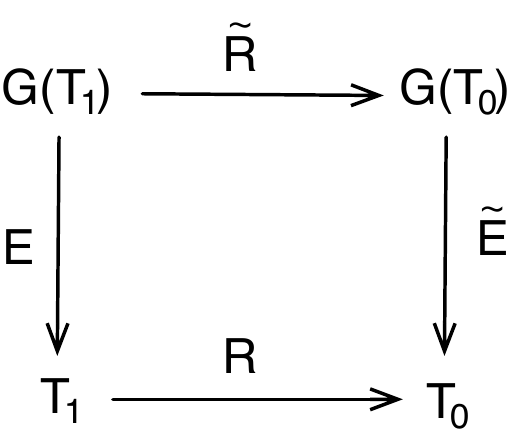} 
      \label{fig:example}
   \end{figure}
   
   Here I have called $E$ the map which effectively links generalizations of $G$ in $T_1$ to sentences of $T_1$. Since there may be only one subset of $T_1$ from which $g$ may be derived (explained), $E$ {\em has to be bijective}.\footnote{This link is mediated by a set called the \underline{product set of T }: $P(T)$, the elements of which are all possible combinations of elements of $T$. This set is the domain of $g$. This more elaborate formulation is in preparation.} 
   \vskip 0.25cm
 We can now state that $R$ preserves nomological necessity iff the diagram {\em commutes}, that is, if there is an {\em unambiguous} way to construct $\tilde E$ (the nomological necessity of $G$ in $T_0$). 
   \vskip 0.25cm
 Now we have seen that $R$ is a map, but it is {\em not} bijective, whereas $E$ {\em is} bijective (a generalization $g$ must have at least one and not more than one explanation in $T_0$). An explanation $\tilde E$ of $g$ in $T_0$ must of course be bijective also. For this to be the case, however, it is a necessary condition that $\tilde R$ must be bijective.\footnote{Since $R$ is not bijective, this is not a sufficient condition. With a few assumptions however one can find sufficient conditions. Since these are not needed here I shall not elaborate on them.} If not, $\tilde E$ will be ambiguous, hence not bijective. This means that no two (or more) generalizations of $g$ in $T_1$ may be mapped onto a generalization of $g$ in $T_0$. This phenomenon however seems to occur when for instance computational psychology is reduced to neurology. The origin of this is the occurrence of ``fusion".\footnote{For a more detailed exposition of fusion, see Fodor (1981b) p. 158ff and references therein.}
   \vskip 0.25cm
   I shall briefly (and sloppily) illustrate fusion:
   \vskip 0.25cm
   As in the previous example, let \\
   
 $x_1=$ ``All elephants are gray"\\
  \mbox{}\ \ \ \ \  $x_2=$ ``Clyde is an elephant"\\

   Since there is a {\em relation of content} between $x_1$ and $x_2$ one may ``infer" the generalization $g=x_1\otimes x_2$, that is $g=$``Clyde is gray" is explained in $T_1$. If now the reduction does not pay attention to content, fusion is likely to occur:
   
  If, for instance: \\
  
 $ y_1=f_1(x_1,x_2)=$``apple"\\
  
$y_2=f_2(x_1,x_2)=$``happy"\\
\vskip 0.25cm
\noindent the generalization ``Clyde is gray" although having a generalization in $T_0$ will {\em not} be nomologically necessary in $T_0$ (i.e. have an explanation in $T_0$), since due to the Reduction $\{f_1,f_2\}$, several generalizations of $g$ in $T_1$ will be translated to {\em one} generalization in $T_0$. What has been distinguishing these generalizations in $T_1$ has been lost in the process of Reduction. I shall call generalizations which are the image of several generalizations {\em degenerated}. I should perhaps emphasize that the term ``fusion" does not refer to the generalizations but to the {\em sentences} of $T_0$: it is obvious that the above Reduction does not preserve syntax and only assigns ``variables" to the new sentences. Such an assignment does not contain any ``pointers" in order to have content. The term ``fusion" refers to the reduction $R$ which fuses expressions of $T_1$ in $T_0$.

Let me now draw the final conclusions of this section which are very similar to Fodor's although the vocabulary I use has been defined much more rigorously.

For a reduction to preserve explanatory power it has to preserve nomological necessity. If fusion occurs, one necessary condition for the Reduction to preserve nomological necessity cannot be met.

With these facts in mind, a cursory look at the reduction of computational psychology to neurology shows that the reducing science will {\em not} have the explanatory power of the reduced science, since in the translated sentences, like ``John is in the mental state XYZ" or ``Susan is in mental state ABC", any hint of content has been lost. These sentences are in effect typically fused expressions, since they contain no referring terms. If psychological statements are actually {\em representations of content}, them the requirement is that this content has to be represented in the reducing science too. I shall now come to the main point of this section.

Although it seems as if the failure of strong Reduction in the computational psychology/neurology case stems from the peculiarity of psychology and its sentences, I shall argue that this is not so. In fact there already exists a system $T_0$ which effectively reduces computational psychology: it is the predicate calculus. Moreover, the predicate calculus is easily translatable into computer-intelligible languages, such as LISP.

\subsection*{Meaning and Content in the Predicate Calculus: How to have Illusions in Illusions}

I shall not be giving a detailed description of the predicate calculus and its applications, since its use is well-known, and much more competent people than I have described and used it at length.\footnote{see for example Charniak (1985)} I shall however give a few striking examples in order to show how the representations of mental states may be related in virtue of their content, and how propositional attitudes fir into the scheme of predicate calculus. Although the predicate calculus may in principle handle sentences on qualitative states (like feelings, emotions, etc.), since there is no psychology of qualitative states I shall give no attention to them. 

As a consequence I shall concentrate only on the representation of the content of propositional attitudes in the predicate calculus and have a look at the structure of the semantical network. For AI-researchers though this section will rather be on the trivial side. 

For convenience I shall give a very short introduction to the predicate calculus, which I have adapted from Charniak, McDermott (1985).
\vskip 0.25cm
An {\em atomic formula} in the predicate calculus is written in the form:
\vskip 0.25cm
(predicate-terms-)
\vskip 0.25cm
as e.g., in
\vskip 0.25cm
\noindent (loves John Mary)\\
(is-gray elephant)\\
(inst block-1 prism)
\vskip 0.25cm
The value of such a bracket is either true or false. They may be connected via the {\em connectives} and, or, not and if, as in:\\
\vskip 0.25cm
\noindent (and (color block-1 yellow)\\
(inst block-1 elephant))\\
\vskip 0.25cm
\noindent which states that block-1 is a yellow elephant. 
\vskip 0.25cm
There are two kinds of {\em quantifiers} in the predicate calculus: the {\em existential} and the {\em universal} quantifiers. 
\vskip 0.25cm
$x$ is a universally quantified variable in formula $f$ with:
\vskip 0.25cm
(forall(x) f).
\vskip 0.25cm
The variable $x$  is existentially quantified with:
\vskip 0.25cm
(exists(x) f).
\vskip 0.25cm
\noindent \underline{Example}:
\vskip 0.25cm
(forall (z) (if (inst z elephant)(color z gray)))  \mbox{}    \hskip 2cm    (3.1)      
\vskip 0.25cm
\noindent expresses the rule that all elephants are gray.
\vskip 0.25cm
Given the last formula, it is possible to infer (color Clyde gray) from the assertion (inst Clyde elephant).  
\vskip 0.25cm
Expression (3.1) is in fact a representation of the belief that elephants are gray. It is in this form that the facts which the computer ``currently" believes are represented. I shall describe in some detail what happens when the computer is asked a simple question, for instance whether the color of Clyde is gray. The main features of the computer which are relevant to us now are the {\em data-base} (or fact-base) and the {\em inference-engine}. The inference-engine is a device which is in possession of inference-rules like (3.1) and accomplishes several tasks, a few of which shall be mentioned subsequently. The data-base on the other hand is crammed with assertions like: (has-part elephant trunk) or: (is-hungry Fred) etc. The inference-engine translates the question into the internal representation (which in this case is the predicate calculus). It then comes up with the command
\vskip 0.25cm
(Show (color Clyde Gray)).

\vskip 0.25cm
It then searches through the data-base in order to prove the ``theorem": (color Clyde gray)). Since it is in possession of rule (3.1), it will immediately come up with (color Clyde gray), the moment it has found the assertion (inst Clyde elephant). In this case, the new fact that (color Clyde gray) may be added to the data-base. The process just described is called ``backward-chaining", since deductions are only carried out if required. There is also another possibility called ``forward-chaining", where the inference engine infers all the assertions it can ``think" of (given the items in the data-base and the rules in the inference-engine). In response to a query the engine searches through the data-base in order to match the query. 

I have been describing this example in some detail to point out just how a relation between generalizations is realized in the predicate calculus. (In order to be brief I have of course been dramatically oversimplifying). Nevertheless it turns out that in effect sentences are pointing to each other in virtue of the ``beliefs" held by the inference engine. The notion of ``chaining" illustrates this quite remarkably. The set of all rules available to the inference-engine links sentences of the data-base into structures like trees and possibly loops. This is what is sometimes called a {\em semantical network}, semantical because the structure stems from relations of content. Now it should be emphasized that the computer, more specifically the inference-engine, has no clue as to what the {\em meaning} is of the symbols it is operating upon, incidentally for the machine the sentences are completely devoid of any content. How this may be possible while we have seen that some sort of content has to be there in order to achieve relations, is in fact straightforward: we are now on a different {\em level of description}\footnote{I borrowed this term from Hofstadter (1979)} than in psychology, and the vocabulary ``meaning" and ``content" is of no use on this level. On the level of the predicate calculus the meaning of a sentence is the background of the semantical network to which it is ``wired". Let me illustrate this with some Wittgensteinian ideas:

In the Wittgensteinian view of philosophy, words are devoid of intrinsic meaning, they acquire their meaning in the background of the possible contexts in which they may be used, that is, the meaning of a word is defined through the way it might be connected to other words. Just {\em how} they may be connected is stipulated by rules to which we abide, the rules of the language-game. The fact that nevertheless we {\em understand} each other is a remnant of our learning a language by rule-following. While as a consequence  the conceptions we have formed of a word are quite similar, the meaning of a word is an intrinsically private matter, indeed one can say that no two persons have exactly the same conception of the meaning of a word, since the ways in which a word is liable to be used depends on the past (verbal or nonverbal) experiences of a person, and these may not be equal. 

It is exactly this way that I understand the meaning of sentences of the predicate calculus. On the level of the machine only, the sentences have no meaning at all. They acquire it by the way they may be connected to other sentences present in the data-base. The {\em rules} that specify just {\em how} to connect them are nothing else than the beliefs held by the inference-engine. Since the structure of the semantical network depends on the facts it contains (remember that any fact inferred as a response to a query may be added to the data-base), the meaning of a sentence (defined as above) may be different for different data-bases.\footnote{The resemblance of the semantical network to neural networks and of the deduction methods like chaining, in the background of the predicate calculus and in neurology, are of course purely coincidental. I shall have nothing to say on neurology.} But on the other hand, due to the fact that the rules of the inference-engine are on a strictly logical basis, the meaning of a sentence like (color Clyde reduce gray) will be quite similar for different data-bases. Let me sum it all up: if a computer is given the possibility to shape its own ``personal" data-base thus creating its ``past", such a computer should be able to produce a near-perfect impression of mind! One last thing that I have promised: there may be in the endless depths of a data-base an innocent-looking {\em sentence} of the form: (have-impression-of mind). This is how the propositional attitude of having-the-impression-of-having-a-mind fits into a theory of the mind without leading to inconsistencies. 

In closing this section, let me briefly comment on an argument brought against this construction by Fodor\footnote{Fodor (1981c)}. I understand his criticism as claiming that:

(a) the representation of meaning in a computer language, and the {\em theory of meaning} associated with sentences in a computer language (called {\em procedural semantics}) is no semantic theory at all, since it does not account for the relation between language and the world.

(b) procedural semantics is a form of verificationism and verificationism is highly implausible. Furthermore it is verificationism that connects language and mind in procedural semantics, thus casting doubt on that connection.

The refutation of these arguments will in fact be purely Wittgensteinian. Sentences of ordinary language have exactly the same relation to the world as have the sentences of predicate calculus. On the level of the sentences {\em alone} (that is, disconnected from their environment, the semantical network), the sentences are entirely meaningless. For example, the sentence ``Boise is a city" creates the illusion of meaningfulness because of the fact that we are unable to disconnect it from its network due to our knowledge of English. On the level of the sentences alone, however, the combination of letters ``Boise" is not referring at all, it is in fact only on the level of the semantical network that Boise refers to a city in Idaho. The same goes for the expression (city Boise). 

The way in which a relation between the meaning of a word
 in ordinary language and an object in the world is established is quite straightforward: it is established through the process of learning a language. In a language we have a word (say: ``book"), and its meaning is given by the ways in which it may be used in different contexts. The relation of the word to an object standing on some shelf, say, has been established by a person pointing to the object and saying the word. The relation is thus created by the process of training in the course of learning a language. This does {\em not} mean that sentences of the language may be mapped onto states of the world, since sentences standing alone do not mean anything. It thus follows, that although sentences of the predicate calculus have truth values, nothing is implied on the states of the world, since the respective notions of truth have nothing to do with each other, being defined on completely disjunct systems. To put it more bluntly: Although (loves John Mary) has a truth value in the predicate calculus, nothing is asserted on the relationship of two people named John and Mary, and yet the sentence (loves John Mary) is meaningful on the level of the semantic network of the predicate calculus. This takes care of the objection concerning verificationism. It is true that on the level of the machine language the structure of the semantical network is lost, and the translated sentences turn meaningless. However, we are not interested in anything on that level. Indeed, machine language is a level of description with typically ``degenerated generalizations", to use the vocabulary of the framework of reduction. But semantical poverty of machine language is not associated with semantical poverty of the higher-level language, as can be seen from the semantical power of the predicate calculus.
 
 \subsection*{Conclusion} 
 
 In this conclusion I shall try to sketch the main line I have been pursuing in the previous section, and subsequently say a few words on the consequences of the structure I have been outlining. 
 
 I have been emphasizing the metaphysical nature of the distinction between real and artificial intelligence and arguing that the only notion that is accessible to us is the impression of intelligence. This was to prepare the ground for discussing this same aspect with respect to the mind. It turned out that in order for the notion of impression of mind to have a meaning one has to be able to demonstrate the possibility of such an impression in an appropriate model of the mind such that the self-referential ``having illusions in illusions" does not lead to paradoxes. Since such a theory of the mind is in general the result of an attempt at reducing psychology, the validity of the procedure, or put in another way, the relevance of the result, has to be investigated for every candidate for a reducer of psychology. As it turns out, the reduction of psychology to computer science, more specifically to the predicate calculus, is satisfactory and sensible, and {\em per constructionem}  enables one to cope with ``illusions in illusions" in a very simple and straightforward way: the propositional attitude of having-the-impression-of having-a-mind is represented as a sentence of the data-base (or, incidentally, as a belief held by the inference-engine). Since this seems like a rather cheap trick, I want to say a few words on the importance of this construction. 
 
 Indeed I think that the fact of having information {\em on} the data-base stored {\em in} the data-base is a far cry from being trivial. I would rather suggest that this feature is one of the essences of our impression of having a mind, in that it is this kind of self-reflective sentences that, (together with another important point to which I shall come soon) convince us that we are dealing with a person, a mind, or whatever have you. What is equally important is the fact that we may provide the computing-system with an additional feature such that these sentences may emerge in the data-base without having been explicitly keyed in. This device is tentatively baptized a ``self-watcher"\footnote{I have borrowed this term from Hofstadter (1985).} and should perform the task of keeping a record of the sentences and the structure of the data-base, and of every information available concerning the data-base itself. When this information is stored away in sentences of the predicate calculus in the data-base itself, the fact that the inference-engine has access to these sentences should produce a curious meshing of sentences on the world with sentences on the data-base which may result in sentences claiming a certain sort of self-awareness. It therefore seems as if the ability of self-reference is indeed the key to the generation of the impression of mind; that consciousness may be nothing but an artifact of a device being able to keep a record of the state of the data-base.
 
 As advertised I want to mention another important thing that plays a crucial role in the impression of mind. It is the observation that people tend to associate intelligent behavior with life in a way that makes it difficult for them to believe that a ``dead object" like a computer may be in a position to produce such a behavior. That this is by no means a trivial point is illustrated by the fact that this has been a major objection to the possibility of AI, namely the claim that mind is intrinsically linked to life, and that a computer is by construction dead and thus as a consequence unable to have a mind (that is, to give the impression of mind). 
 
 I want to argue against this point by making use of the change of perspective that is at the root of this essay. Since I have been emphasizing this perspective twice already, I guess that I may be brief on that point.
 
 The criterion for what constitutes the essence of life should be the impression of life, there should be no metaphysical distinction between living and dead. Let me illustrate this by the following. When an allegedly living creature is viewed from different levels of description, one notices that if one goes down subsequent levels, say from the body to the parts of the body, to the organs, the cells, the cell organelles, the molecules, the atoms, and eventually to the elementary particles, what we generally call life suddenly disappears. What then, in the light of this fact, may be the essence of life? Surely nothing metaphysical, since this should  then be present at every level.\footnote{It is hardly conceivable that a metaphysical quality only pertains to sone arrangement of objects and not to the objects themselves, or put in a fancier way: a metaphysical quality is a metaphysical quality.} What may change is the impression of life at different levels due to the emergence of collective phenomena, the origin of which lie in the dynamics of the objects present at the underlying levels to which one has no access at the level of the collective phenomena. No object should therefore be treated as either alive or dead, since these are no global categories. They have been promoted to global categories in ordinary language, which gave rise to such concepts as the soul, which then plays the role of the metaphysical distinction between living and dead. This in turn triggers problems like what happens to the soul when the creature eventually dies. The moment one drops the metaphysical distinction between life and death such problems are absent since it is conceivable that a system stops to give the impression of life. 
 
 This perspective obviously takes care of the objection mentioned earlier in that it makes no sense to say that a computer is not alive. All one can say is that it fails to give the impression of life, but that may be changed with the advent of more and more clever programs, and is not an insurmountable barrier. Maybe one should start with systems whose data-bases may not be erased, in that effect allowing the system to shape its``personal" data-base thus effectively shaping its ``language-game" .
 
 It is hardly conceivable that this is what Wittgenstein had in mind when he used ``language-game" and ``form-of-life" synonymously, but the view I have developed in this essay at least gives some hint on how it was possible that his way of putting things so often was superior to nonanalytical views of philosophy.
\newpage
\section*{References}

\mbox{}\hskip 0.6cm Cherniak, E., Mc Dermott, D.V. 1985. {\em Artificial Intelligence}. Addison Wesley.

Descartes, R. 1637. {\em Discourse on Method}. In Haldane and Ross, 1968.

Descartes, R. 1641. {\em Meditations}. In Haldane and Ross, 1968.

Flanagan, O.J. 1984, {\em The Science of the Mind}. Cambridge: the MIT Press/Bradford Books.

Fodor, J. 1981. {\em RePresentations: Essays on the Foundations of Cognitive Science}. Cambridge: The MIT Press/Bradford Books.

Fodor, J. 1981a. ``Operationalism and Ordinary language". reprinted in Fodor, 1981.

Fodor, J. 1981b. ``Computation and Reduction". reprinted in Fodor 1981.

Fodor, J. 1981c. ``Tom Swift and the Procedural Grandmother". reprinted in Fodor 1981.

Haugeland, J. 1985. {\em Artificial Intelligence: The Very Idea}. Cambridge: the MIT Press/Bradford Books.

Hofstadter, D.R. 1979. {\em G\"odel, Escher, Bach: An Eternal Golden Braid.} New York: Vintage Books. 

Hofstadter, D.R. 1985. {\em Metamagical Themas}. New York: Basic Books. 

Lang, S. 1970. {\em Linear Algebra}. Addison Wesley.

Rorty, R. 1979. {\em Philosophy and the Mirror of Nature}. Princeton: Princeton University Press. 

Wittgenstein, L. {\em \"Uber Gewi{\ss}heit. On Certainty. G.E.M. Anscombe, G.H. von Wright, eds. Oxford: Basil Blackwell.

\bibliographystyle{plain}
\bibliography{encyc}
\end{document}